# Graph-Grammar Assistance
# for Automated Generation of Influence Diagrams

**John W. Egar and Mark A. Musen**
Section on Medical Informatics
Stanford University School of Medicine
Stanford, CA 94305-5479
*email:* {egar, musen}@camis.stanford.edu

## Abstract

One of the most difficult aspects of modeling complex dilemmas in decision-analytic terms is composing a diagram of relevance relations from a set of domain concepts. Decision models in domains such as medicine, however, exhibit certain prototypical patterns that can guide the modeling process. Medical concepts can be classified according to semantic types that have characteristic positions and typical roles in an influence-diagram model. We have developed a graph-grammar production system that uses such inherent interrelationships among medical terms to facilitate the modeling of medical decisions.

## 1  MODELING OF DECISIONS

Where judgments are made under appreciable uncertainty, and where stakes are high, decision makers may need help in weighing risks and benefits involved in important decisions. Decision theory can provide such help. Unfortunately, constructing decision-analytic models is a difficult task, even for trained analysts.

We have implemented a **graph-grammar production system** that constructs decision-analytic models automatically from unordered lists of standard terms. A **graph grammar** is a system of replacement rules that operates on graphs, rather than on the strings on which traditional string grammars operate. Our current graph grammar manipulates influence-diagram graphs to incorporate concerns that a user lists for a particular decision problem.

The decisions that we have been modeling involve selecting an optimal plan for medical intervention. The plan may consist of tests, treatments, and specimen collections, and is tailored to individual patients, each of whom has her own particular set of priorities and her individual probability distributions for the various chance outcomes. Since each patient must make a decision for her particular circumstances, we need to tailor a model to fit each dilemma so that the model can properly guide the patient, or some health-care worker acting for the patient, toward a normative decision. Our system accepts a list of patient concerns, and generates a qualitative decision model that includes those concerns as variables. Decision analysts must then perform subsequent assessment of quantitative probabilities and utilities for the generated model.

## 2  GRAPH GRAMMARS

A graph grammar consists of a set of production rules that dictates how a graph can be transformed and rewritten. These production rules are quite different from the productions used in rule-based expert systems: Graph-grammar rules can specify a wide range of contexts for which they are applicable, and can describe different graph manipulations for those different contexts. A graph grammar specifies a language over a set of symbols, the members of which are elements of a graph. We have found graph grammars to offer an expressive and concise way to represent prototypical forms for modeling dilemmas. Also, graph grammars can provide high-level abstractions that help users to manage complexities.

Over the past 20 years, researchers have developed a plethora of formalisms to describe rewriting procedures for graphs. These formalisms include algebraic graph grammars, array grammars, collage grammars, edge-label–controlled grammars, expression grammars, graphic sequential rewriting systems, hyperedge-replacement grammars, map grammars, neighborhood-controlled embedding grammars, node-label–controlled grammars, picture-layout grammars, plex grammars, precedence graph grammars, relation grammars, shape grammars, and web grammars (Ehrig et al., 1991; Pavlidis, 1972). The particular formalism that we use is a modification of Göttler's operational graph grammars (Barthelmann, 1991).

Each **production rule** in a grammar describes a le-gal graph manipulation. In Göttler's formalism, we write these **productions** as graphs divided into four **regions** (Figure 1), which partition the vertices into four sets: those in the left region, $V_L$; those in the right region, $V_R$; those in the **indeterminate** region above, $V_A$; and those in the **determinate** region below, $V_B$. The two sets $V_A$ and $V_B$ are referred to as the **embedding part.** All vertices ($V$) in the production and the host graph have labels ($L_V$) and a mapping ($l_V : V \rightarrow L_V$) from vertices to their labels. There is a finite set of edge labels ($L_E$) for the directed edges ($E \subseteq V \times V \times L_E$) of productions and host graphs. The graph manipulation described by such a production is as follows: Find nodes matching the left region, $V_L$, and replace them with nodes matching the right region, $V_R$. The procedure consists of these four steps:

1. Find a region of the host graph where the vertices and edges match the vertices and edges of the determinate ($V_B$) and left ($V_L$) regions of the production (Figure 2a).[1]

2. Find zero or more edges that match edges between the left ($V_L$) and indeterminate ($V_A$) regions of the production.[2]

3. Remove from the host graph those vertices that matched vertices within $V_L$ (Figure 2b).

4. Add to the host graph new vertices and edges that correspond to those within the right region ($V_R$) of the production, and add to the graph edges that correspond to those connecting the embedding part (from $V_B$ and the matched portion of $V_A$) of the production to the right region ($V_R$) (Figure 2c).

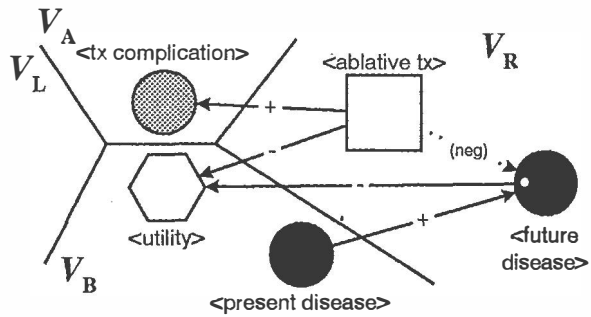

Figure 1: Sample graph-grammar production rule. This production rule describes how nodes of the type <ablative tx> can be added to the host graph. $V_L$, $V_A$, $V_B$, and $V_R$ (left, above, below, and right) are the four regions of a graph-grammar production rule. In this rule, there are no vertices in $V_L$. (tx = treatment)

---

[1] If $V_L$ and $V_B$ match multiple subgraphs, the user must select the appropriate match.

[2] The user of our system must confirm or reject any potential matches to $V_A$. Edges between $V_L$ and $V_R$ or between $V_A$ and $V_B$ are not permitted.

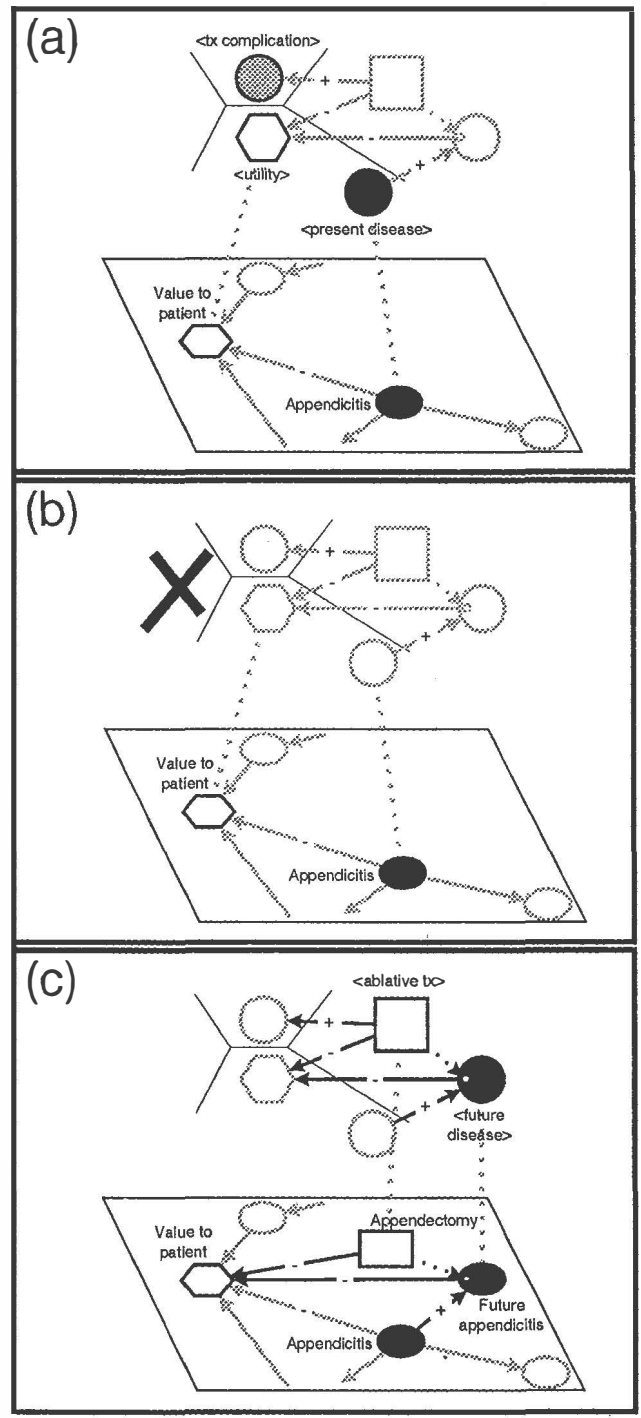

Figure 2: Sample application of the graph-grammar rule from Figure 1. (a) The first view of the host graph shows two nodes from the host diagram matching <utility> and <present disease> in the production. (b) If $V_L$ contained vertices, a matching set of nodes would be removed. (c) Additional nodes Appendectomy and Future appendicitis are added to the QCID model.

Our graph-grammar production system requires a set of such replacement rules. (For details of the graph-grammar formalism and the behavior of our production system, see Appendix A.) Often, where there is simple node replacement with no constraints on the embedding environment, we shall abbreviate such productions with string-grammar productions.

We have implemented a graph-grammar derivation system, called Gramarye, which follows precisely this formalism. The derivation system runs under Common Lisp. We have also implemented a NeXTSTEP user interface for Gramarye.

## 3   A GRAPH GRAMMAR FOR MEDICAL DECISIONS

Gramarye must be given four pieces of static information before it generates any models:

1. A collection of graph-grammar productions

2. An initial graph

3. A classification of node labels according to abstract symbols used in the productions

4. A visual notation for each subclass of node

The particular application with which we are concerned in this work is that of generating medical decision models. Consequently, to construct influence diagrams from medical terms, we use the following static input:

1. A graph grammar for medical influence diagrams

2. The utility node, Value to patient

3. A classification tree for a medical lexicon

4. The shapes rectangle, circle, and hexagon for decision, chance, and utility nodes, respectively

Currently, Gramarye consists of a user interface for loading the static input, a command-line interface for the accepting terms and for interacting with the user during the derivation process, hash tables for storing the vocabulary and the graph grammar, the derivation mechanism, a layout algorithm, and a diagram generator.

Our present underlying graph representation—**qualitative contingent influence diagram**, or **QCID**—is an extension to the **influence-diagram** notation (Howard and Matheson, 1984). We use **contingent nodes,** a notation described by Fung and Shachter (Fung and Shachter, 1990). In this notation, we divide a node into several contingent nodes, each with exclusive conditions. Each contingency node is considered relevant to the rest of the diagram for only those scenarios in which its conditions are met. Also, following Wellman (1990), we represent qualitative relationships by labels on the probabilistic arcs: A plus sign, "+," indicates a direct relationship; a minus sign, "−," indicates an inverse relationship; a question mark, "?," indicates an unclear or nonmonotonic relationship.

The current graph grammar appears in Figure 3. All node labels in the grammar are abstract classes for standard medical terms. The graphical representation helps the developer of the grammar to identify structural motifs. Although neither Gramarye nor the QCID notation assign any significance to node coloration, we have colored diagnostic nodes black and other chance nodes light gray to make the grammar's structural patterns more apparent. All the information in Gramarye regarding prototypical patterns in medical decision models can be viewed as 14 rules, each with no more than seven nonterminal symbols.

One of us (Egar, 1993) has shown that the graph grammar that Gramarye currently uses maintains the following properties in all derived QCID models:

1. The directed graph is acyclic.

2. There are no qualitatively dominated decision nodes.

3. At most one derivation can result from a given input[3] (i.e., the grammar is unambiguous).

4. There is exactly one overall utility node.

5. There are no successors to the utility node.

6. All nodes have some path to the utility node.

7. All chance nodes have paths to the value node with no intervening decision nodes.

The second property—no dominated decisions—accounts for five of the nine rules[4] used by Wellman and colleagues to critique manually composed decision trees (Wellman et al., 1989).

Note that the grammar is not in a stable state: We anticipate many changes. One possible change would be to expand the grammar and the vocabulary to make more strictly semantic distinctions. For example, the current grammar does not distinguish among terms that pertain to different organ systems and different clinical domains. A more elaborate classification hierarchy might group electroencephalogram (EEG) and electrocardiogram (EKG) tests in separate nonterminal categories, and a more elaborate graph grammar might then restrict EEG tests to neurologic diseases, and EKG tests to cardiac diseases.

The classification hierarchy of medical terminology gives Gramarye the ability to assign each entered term to one or more structural patterns. If the user enters a term that is not already classified according to the nonterminals of the graph grammar, then she must classify the term manually. Also, the current vocabulary is fairly limited: The vocabulary contains roughly 5000 findings, 850 diseases, and 85 other terminal and nonterminal symbols. Most of the terms for findings and diseases were derived from the terminology used by the QMR system (Shwe et al., 1991).

---

[3] Here, we use the term *input* to refer both to the initial list of concerns and to the choices made by the user when multiple matches to $V_A$ are present.

[4] The other four critiquing rules are obviated by use of an influence-diagram notation.

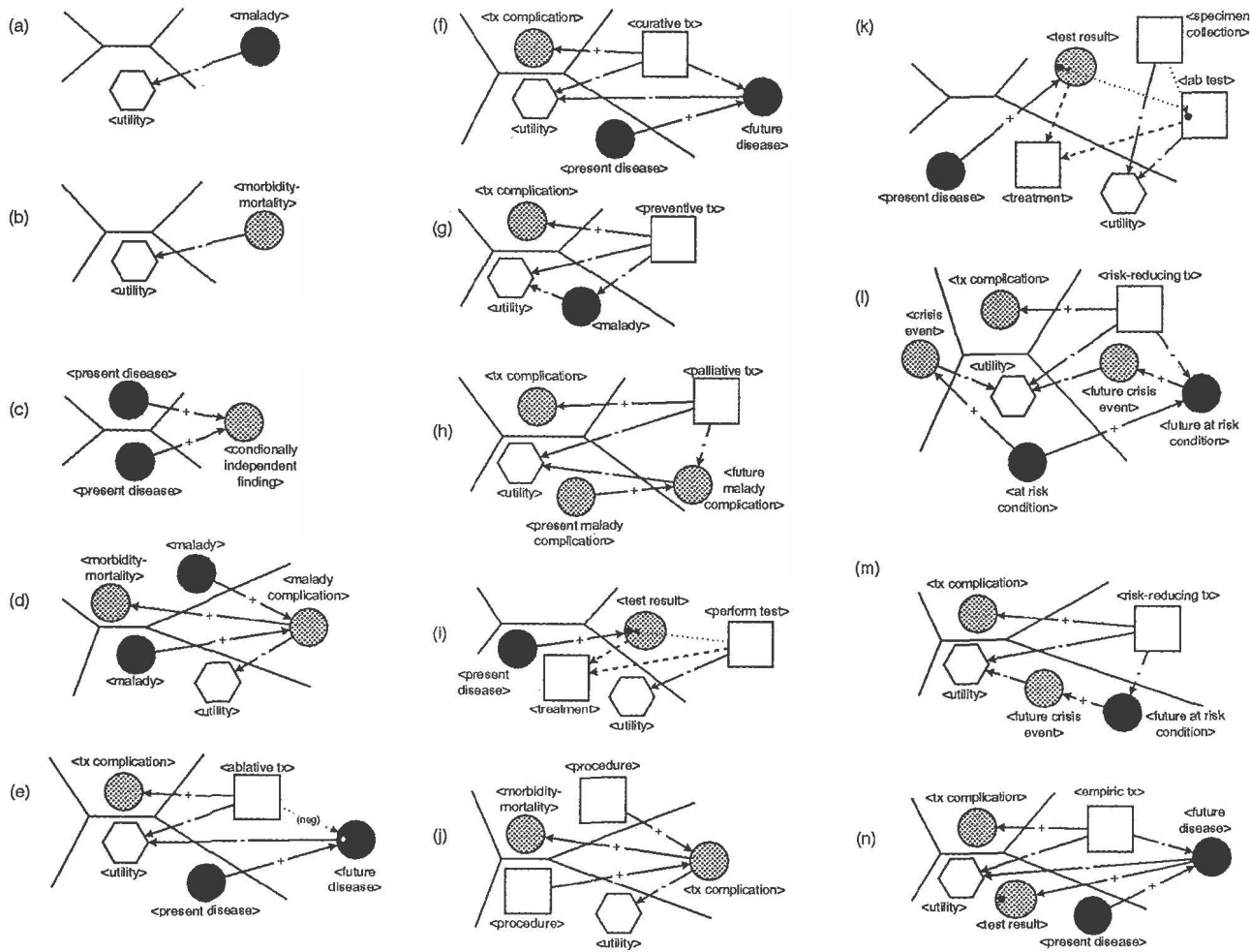

Figure 3: The current graph grammar used by Gramarye to derive medical QCID decision models. These rules add nodes that represent medical concepts from the following categories: (a) maladies, (b) morbidities or mortality, (c) conditionally independent findings, (d) complications of one or more maladies, (e) ablative treatments, (f) curative treatments, (g) preventive treatments, (h) palliative treatments, (i) tests, (j) treatment complications, (k) laboratory tests that require the collection of some specimen, (l) risk-reducing treatments, (m) subsequent risk-reducing treatments, and (n) empiric treatments. For visual clarity, we have colored disease nodes black, and other chance nodes light gray. (tx = treatment)

We have used Gramarye and the grammar described to derive several medical decision models, including a close match to a qualitative influence diagram with more than 20 nodes described in the literature (Wellman, 1990). In models where multiple nodes of the same type are present, the derivation system may require user assistance to choose how subsequent nodes are added to the evolving graph.

Although the grammar in Figure 3 is medically oriented, Shachter has rewritten these rules as a smaller and more general QCID grammar that contains only nonmedical abstractions.[5] Consequently, both medical and nonmedical QCID models may follow a syntax which is more comprehensive than the one we are using.

## 4    DISCUSSION

Several investigators have found that knowledge bases can imply a directed network of causality that can be interpreted as a belief network (Breese, 1992; Leong, 1992; Wellman et al., 1992; Horsch and Poole, 1990; Laskey, 1990; Provan and Clarke, 1993). Goldman and Charniak (1990) use a system of rules that describe how to transform generic relations into probabilistic arcs, and how to expand the conditional probability matrices at the tail of the newly added arcs. All these approaches rely on an external process to provide relationships among the specific variables that appear in the generated influence diagram. In contrast, we rely on a graphical syntax and a semantic classification for the variable labels to infer the relationships among the variables.

---

[5]Ross D. Shachter. Personal communication.

Researchers have modeled a set of decisions associated with a specific medical subject, such as infertility (Holtzman, 1988), and then have pruned the influence-diagram structure to leave only those considerations that are pertinent to an individual case. This approach requires that there be an exhaustive model that is roughly the graph union of all the decision models for a particular subject. On a grander scale, Heckerman and Horvitz (1990) have proposed a comprehensive decision model to include all of internal medicine, based on the QMR diagnostic model (Shwe et al., 1991).

The strength of our approach lies in our ability to change a constructive task—where the user must decide to include or omit each of a large number of possible arcs—to a classification task. In the latter case, the difficulty of our modeling task grows linearly with the number of considerations to be included in the model. The weaknesses that are evident in the current system fall into three categories: (1) Gramarye's inability to include large chance-node subgraphs; (2) its demand for user assistance when a rule contains nodes in the indeterminate region, or when a rule can be applied at several locations in the host graph; and (3) its small vocabulary. Although changes to the graph grammar may correct a few of these deficiencies, we suspect that additional domain information and a more sophisticated node-label classification hierarchy are required for significant improvements in Gramarye's performance.

### 4.1   INCLUSION OF DOMAIN-SPECIFIC INFORMATION

The quality of the derived models is affected adversely by the grammar's complete ignorance of physiologic relationships. The grammar's implicit assumption that all findings are conditionally independent is often unwarranted. Also, the lack of complete ordering for the decision nodes must be rectified by the user for most complex models. Furthermore, when *clusters* of chance nodes are pertinent to the decision problem, it seems that a domain-specific model of medical relationships—beyond the model implied in our graph grammar—is required for the automatic derivation of these qualitative models. In short, some concerns in a decision problem can be modeled in a typical structure; other concerns require a specific understanding of their causal and correlational relationships before they can be modeled into an influence diagram. Whether adequate knowledge bases and belief networks exist that might supply node clusters for the latter concerns is a subject of our ongoing investigation.

Because the assistance required for large models can be annoying to the user, and because the current grammar assumes that findings are conditionally independent given the diseases to which they are relevant, we are investigating how existing belief networks can provide domain-specific relevance information. In particular, we are using the QMR-BN (Shwe et al., 1991) network of diagnostic medicine to guide the derivation and to provide additional arcs that would otherwise be absent from the derived model. One problem that we have faced is the lack of a standardized vocabulary used in existing belief-network models.

### 4.2   EXPANSION OF THE NODE-LABEL CLASSIFICATION

Clearly, as it now stands, Gramarye's vocabulary is not sufficiently comprehensive for general use. We see two possible—and by no means exclusive—solutions to this problem: (1) we can increase the size of the vocabulary to cover most foreseeable entries, and (2) we can limit the user's entries to codes provided by the computer.

The first solution is feasible with only large, standardized, structured clinical lexicons, such as the SNOMED-III (Rothwell and Coté, 1990), or UMLS (Tuttle et al., 1992). The UMLS vocabulary contains over 240,000 terms; where synonyms exist, a single preferred term is associated with each synonym. The terms in UMLS are already grouped according to a detailed classification tree, and according to a network of semantic relations. SNOMED-III uses many of the same semantic types found in our grammar to group over 200,000 standardized terms. A standard vocabulary provides the additional benefit of a well-defined referential semantics: Each term can be associated with a precise English explanation, so that there is little confusion among users regarding what the term means in the real world. Such semantics might enhance the shareability of assessed probabilities and default utilities. However, many decision problems faced by physicians and patients include nonmedical considerations that cannot be pre-enumerated with present technology, so a predefined vocabulary will almost certainly not be sufficient for all clinical decision problems. Also, while the terms included may cover most medical concepts, the user would probably use words and phrases outside the specific lexicon.

The second solution to the restrictions of a single finite vocabulary—limiting entries to those from a computer—could involve an electronic medical record that produces a coded representation from a friendly user interface. Campbell and Musen (1992) have developed one possible user interface, along with a plan for a formal representation for the patient record. In the system that they envision, a semantic network based on SNOMED-III stores all progress-note information. This type of graphical representation might be used by a system such as Gramarye, not only for the unordered list of terms, but also for additional context that might disambiguate how a decision model should be constructed.

In summary, we have found that our graph-grammar production system facilitates the modeling of medical dilemmas. Graph grammars address relationships among medical concepts other than lexical ordering; consequently, graph grammars are ideally suited for deriving a decision model from an unordered list of medical concerns.

## Acknowledgments

We are most grateful to John Breese, Keith Campbell, Henrik Eriksson, David Heckerman, Ross Shachter, Edward Shortliffe, and Jaap Suermondt for their support and guidance. We thank three anonymous reviewers for their helpful comments. We are especially grateful to Lyn Dupré for editing a previous draft of this paper.

We conducted this work with the support of the National Library of Medicine under grants LM-05157 and LM-07033. Computing support was provided by the CAMIS resource, funded under grant number LM-05305 from the National Library of Medicine.

# A GÖTTLER'S GRAPH-GRAMMAR FORMALISM

In the more formal terms of Barthelmann's notation (1991), we define a labeled directed graph as comprising three sets:

1. A set of vertices ($V$), with labels ($L_V$) and a mapping ($l_V : V \to L_V$) from vertices to their labels
2. A set of permissible edge labels ($L_E$)
3. A set of labeled directed edges ($E \subseteq V \times V \times L_E$)

## A.1 CONNECTED VERTICES

A spanned subgraph, span($V', G$), of the host graph $G$ and spanning vertices $V'$ consists of vertices ($V' \subseteq V_G$), their labels ($L_{V_G}$), the edges between vertices in $V'$ ($V' \times V' \times L_{V_G} \cap E_G$), their labels ($L_{E_G}$), and the restricted labeling function $l_V(V')$. A **chain** is a sequence of edges, without regard to their direction. So, a chain between $v_0$ and $v_N$ exists if and only if

$$(\forall i \in \{1, \dots, n\}) \quad (\exists m \in L_E)((v_{i-1}, v_i, m) \in E \vee (v_i, v_{i-1}, m) \in E). \tag{1}$$

For a given node $v$, and for a given set of vertices $V_X$, all the nodes $v'$ that are reachable by some chain through nodes in $V_X$ are considered **connected** to $v$, and this relation is denoted $v \sim v'$. We shall use the connection relation to determine which of the vertices in the indeterminate region are to be mapped to the host graph, where we shall add edges as specified in the production.

## A.2 APPLICABILITY OF A PRODUCTION

If all vertices in the left and bottom regions of the graph-grammar production are matched to vertices in the host graph, and if the edges among the vertices in the left and bottom regions of the production are matched to corresponding edges in the host graph, then we can apply the rule. However, we must note which nodes and edges in the host graph match those in the indeterminate region and are connected, by the reflexive $\sim$ relation described in Section A.1, to the nodes and edges that we have already matched to the left and bottom regions of the production.

So, more formally, a production rule $p$ is **applicable** to some subgraph of $G$ indicated by the monomorphism $\delta : \text{span}(V_{L,p} \cup V_{B,p}, p) \to G$ if and only if

$$
\begin{aligned}
(\forall v \in V_{L,p})(\forall(\delta(v), v', m) \in E_G) \\
(\delta(v), v', m) \in \delta(E_p \cap V_{L,p} \times V_{L,p} \times L_E) \\
\cup \bigcup_{C \in V_{A,p}/\sim} \bigcup_{\mu \in M_\delta(C)} \text{Old}(\mu),
\end{aligned}
\tag{2}
$$

and

$$
\begin{aligned}
(\forall v \in V_{L,p})(\forall(v', \delta(v), m) \in E_G) \\
(v', \delta(v), m) \in \delta(E_p \cap V_{L,p} \times V_{L,p} \times L_E) \\
\cup \bigcup_{C \in V_{A,p}/\sim} \bigcup_{\mu \in M_\delta(C)} \text{Old}(\mu),
\end{aligned}
\tag{3}
$$

where

$$
\begin{aligned}
M_\delta(C) &= \{\mu : \text{span}(V_{L,p} \cup V_{B,p} \cup C, p) \\
&\quad \to G | \mu(\text{span}(V_{L,p} \cup V_{B,p})) = \delta\} \quad (4) \\
\text{Old}(\mu) &= \{(\delta(v), \mu(v'), m) \mid (v, v', m) \in E_p \\
&\quad \cap V_{L,p} \times (V_{A,p} \cup V_{B,p}) \times L_E\} \\
&\quad \cup \{(\mu(v'), \delta(v), m) \mid (v', v, m) \in E_p \\
&\quad \cap (V_{A,p} \cup V_{B,p}) \times V_{L,p} \times L_E\}. \quad (5)
\end{aligned}
$$

The set of edges $\text{Old}(\mu)$ contains those edges between the vertices to be removed and the embedding environment in the host graph. The set $C$ represents connected vertices in $V_A$. The set of edges $M_\delta(C)$ contains those edges in the host graph that are matched to those in $V_A$ according to $\mu$—a particular (indeterminate) extension of the vertex monomorphism, $\delta$. The edges in the set $\bigcup_{C \in V_{A,p}/\sim} \bigcup_{\mu \in M_\delta(C)} \text{Old}(\mu)$ are those edges that connect vertices to be removed with vertices in matched subgraphs in their $\sim$-connected environment, along with edges in that matched subgraph. So, all host-graph edges incident to the vertices that match those in $V_L$, according to $\delta$, can be divided into

1. Edges that match edges in $V_L$
2. Edges that match edges (including matched target subgraphs) from $V_L$ to the embedding environment
3. Edges that do not match edges in the production

We must distinguish the target subgraphs of the second group of edges, because they are precisely those subgraphs that may need to be connected to the inserted subgraph ($V_R$) according to the production.

## A.3   EFFECTS OF A PRODUCTION

When a production $p$ is applied to the host graph, there are three basic effects:

1. We remove labeled vertices matching those in $V_{\mathrm{L}}$.
2. We add labeled vertices corresponding to those in $V_{\mathrm{R}}$.
3. We add labeled edges among the new nodes and between the new nodes and subgraphs of the host that were connected to the removed vertices and matched according to a particular extension of $\delta$, all according to the production.

Stated formally, the effect of applying production $p$ on graph $G$ at the subgraph indicated by $\delta : \mathrm{span}(V_{\mathrm{L},p} \cup V_{\mathrm{B},p}, p) \to G$ is the graph $H$:

$$V_H = V_G \setminus \delta(V_{\mathrm{L},p}) \cup V_{\mathrm{R},p} \qquad (6)$$

$$l_{V_H} = l_{V_G}(V_G \setminus \delta(V_{\mathrm{L},p})) \cup l_{V_p}(V_{\mathrm{R},p}) \qquad (7)$$

$$\begin{aligned} E_H = & (E_G \cap (V_G \setminus \delta(V_{\mathrm{L},p})) \times (V_G \setminus \delta(V_{\mathrm{L},p})) \times L_E) \\ & \cup (E_p \cap V_{\mathrm{R},p} \times V_{\mathrm{R},p} \times L_E) \\ & \cup (\bigcup_{C \in V_{\mathrm{A},p}/\sim} \bigcup_{\mu \in M_\delta(C)} \mathrm{New}(\mu)), \end{aligned} \qquad (8)$$

where

$$\begin{aligned} M_\delta(C) = & \{\mu : \mathrm{span}(V_{\mathrm{L},p} \cup V_{\mathrm{B},p} \cup C, p) \\ & \to G | \mu(\mathrm{span}(V_{\mathrm{L},p} \cup V_{\mathrm{B},p})) = \delta\} \qquad (9) \\ \mathrm{New}(\mu) = & \{(\mu(v), v', m)|(v, v', m) \in E_p \\ & \cap (V_{\mathrm{A},p} \cup V_{\mathrm{B},p}) \times V_{\mathrm{R},p} \times L_E\} \\ & \cup \{(v', \mu(v), m)|(v', v, m) \in E_p \\ & \cap V_{\mathrm{R},p} \times (V_{\mathrm{A},p} \cup V_{\mathrm{B},p}) \times L_E\}, \quad (10) \end{aligned}$$

and where the labeling set does not change:

$$\begin{aligned} L_{V_H} &= L_V, \\ L_{E_H} &= L_E. \end{aligned}$$

The edges $\bigcup_{C \in V_{\mathrm{A},p}/\sim} \bigcup_{\mu \in M_\delta(C)} \mathrm{New}(\mu)$ are the edges that connect vertices to be added with the matched subgraph in the connected environment, along with edges in the subgraph that is matched to $V_{\mathrm{A}}$.

## A.4   CONSTRUCTIVE DERIVATION OF GRAPHS

To make Göttler's formalism convenient for modeling medical decisions, we have adopted four modifications:

1. When the application of a production cannot be determined from the host graph, the monomorphism and the set $C$ are determined by a derivation system's user.
2. The labeling function also matches nonterminal symbols (denoted by angle brackets) in a production to terminal symbols (instances) in a classification hierarchy.

3. The derivation system synthesizes variant symbols from nodes already matched in a monomorphism, and adds the new terminal symbols to the classification hierarchy.
4. Those vertices in $V_{\mathrm{R}}$ that match nodes that already exist in the host graph are not duplicated (Equations 6 and 7).

The latter conditions allow for accessory variant nodes ($V_{\mathrm{R}_{\mathrm{var}},p}$) matching those in $V_{\mathrm{R}}$ to be added as needed. We partition vertices in the right region of a production into two sets: $V_{\mathrm{R}_{\mathrm{new}}}$ and $V_{\mathrm{R}_{\mathrm{var}}}$. Those vertices ($V_{\mathrm{R}_{\mathrm{var}}}$) with nonterminal labels that are marked as variant labels in the node-label classification hierarchy are matched to new nodes with labels that consist of the indicated prefix or suffix and a stem that is identical to the label string of another matched node in the production.[6] When a variant node, derived from a node already matched in the monomorphism, is identical to a node in the host graph, it is replaced by the host-graph node in that monomorphism. We define a **derivation** ($\mathcal{D}$) to be a sequence of **partial derivation stages** ($\delta_i$). Each partial derivation stage comprises a set of monomorphisms ($\delta_i$) of various—not necessarily distinct—productions that are applied to the host graph in an arbitrary order. Each monomorphism must introduce at least one node with a label from the input list ($L_{V_{\mathrm{input}}}$) to the host graph. We also require that each application in a particular stage must (1) introduce a new node to the host graph, and (2) be applicable to the host graph prior to any modifications made by other applications of that derivation stage. That is, $\delta_i$ consists of all possible applications of productions such that each adds a different label from $L_{V_{\mathrm{input}}}$ to the host graph, and such that the following properties hold for each application $\delta$ :

$$V_{\mathrm{R}_{\mathrm{new}},p} \cap V_G = \emptyset, \qquad (11)$$

$$l_{V_H}(V_{\mathrm{R}_{\mathrm{new}},p}) \in L_{V_{\mathrm{input}}}, \qquad (12)$$

$$V_{\mathrm{R}_{\mathrm{new}},p} \subseteq V_H, \qquad (13)$$

where $G$ is the host graph before $\delta_i$, and $H$ is the host graph after $\delta_i$. A derivation $\mathcal{D}$ is successful if the host graph resulting from the last derivation stage contains a node corresponding to each term in the input list, $L_{V_{\mathrm{input}}}$. A grammar is **ambiguous** if multiple graphs can be derived from the same input. Here, *input* refers to both the list of terms, $L_{V_{\mathrm{input}}}$, and the user's responses to $V_{\mathrm{A}}$-matching queries from the system.

---

[6] When there are zero or more than one vertex of a sibling or parent class to a variant label, additional notation will be needed to resolve this ambiguity. Our grammar has not yet required such extensions to the graph-grammar formalism.